\documentclass{IEEEcsmag}

\usepackage[colorlinks,urlcolor=blue,linkcolor=blue,citecolor=blue]{hyperref}
\newcommand{\upcite}[1]{\textsuperscript{\textsuperscript{\cite{#1}}}}
\usepackage[numbers,sort&compress]{natbib} \setcitestyle{open={},close={}}

\expandafter\def\expandafter\UrlBreaks\expandafter{\UrlBreaks\do\/\do\*\do\-\do\~\do\'\do\"\do\-}
\usepackage{upmath,color}

\usepackage{array}
\usepackage{url}
\usepackage{verbatim}
\usepackage{graphicx}
\usepackage{multirow}
\usepackage{makecell}
\usepackage{amsthm}
\usepackage{amsmath,amsfonts}
\usepackage{algorithmic} 
\usepackage[ruled, vlined]{algorithm2e}

\newcommand{\eg}{\textit{e.g. }}
\newcommand{\al}{\textit{et al. }}
\SetKwInput{KwInput}{Input}
\SetKwInput{KwOutput}{Output}

\usepackage{textcomp}
\usepackage{stfloats}
\usepackage{booktabs}
\graphicspath{{fig/}}

\jvol{XX}
\jnum{XX}
\paper{8}
\jmonth{Month}
\jname{Publication Name}
\jtitle{Publication Title}
\pubyear{2023}

\begin{document}


\title{Group Benefits Instances Selection for Data Purification}

\author{Zhenhuang~Cai}
\affil{Nanjing University of Science and Technology, Nanjing, 210014, China}

\author{Chuanyi~Zhang}
\affil{Hohai University, Nanjing, 210024, China}

\author{Dan~Huang}
\affil{Beijing Institute of Technology, Beijing, 100081, China}

\author{Yuanbo~Chen}
\affil{Beijing Research Institute of Mechanical and Electrical Technology, Beijing, 100083, China}

\author{Xiuyun~Guan}
\affil{China Ordnance Industrial Standardization Research Institute, Beijing, 100089, China}

\author{Yazhou~Yao{$^*$}}

\affil{Nanjing University of Science and Technology, Nanjing, 210014, China}

\markboth{THEME/FEATURE/DEPARTMENT}{THEME/FEATURE/DEPARTMENT}

\begin{abstract}	
Manually annotating datasets for training deep models is very labor-intensive and time-consuming. To overcome such inferiority, directly leveraging web images to conduct training data becomes a natural choice. Nevertheless, the presence of label noise in web data usually degrades the model performance.
Existing methods for combating label noise are typically designed and tested on synthetic noisy datasets. However, they tend to fail to achieve satisfying results on real-world noisy datasets. To this end, we propose a method named GRIP to alleviate the noisy label problem for both synthetic and real-world datasets. Specifically, GRIP utilizes a group regularization strategy that estimates class soft labels to improve noise-robustness. Soft label supervision reduces overfitting on noisy labels and learns inter-class similarities to benefit classification. Furthermore, an instance purification operation globally identifies noisy labels by measuring the difference between each training sample and its class soft label.
Through operations at both group and instance levels, our approach integrates the advantages of noise-robust and noise-cleaning methods and remarkably alleviates the performance degradation caused by noisy labels. Comprehensive experimental results on synthetic and real-world datasets demonstrate the superiority of GRIP over the existing state-of-the-art methods. The data and source code of this work have been made available at: {\url{https://github.com/NUST-Machine-Intelligence-Laboratory/GRIP}}.
\end{abstract}

\maketitle

\begin{figure}[t]
	\centering
	\includegraphics[width=0.45\textwidth]{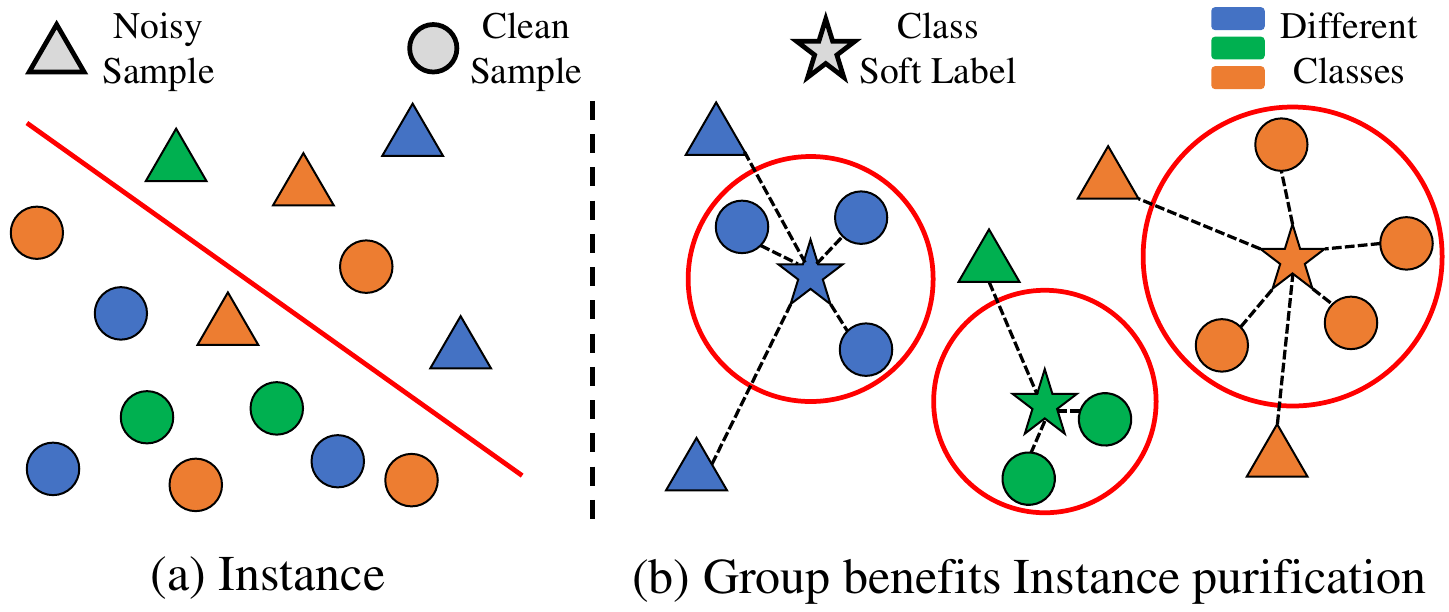}
	\caption{Our approach (b) boosts the typical noise identification (a) through a group regularization strategy. Specifically, it utilizes the similarity between the predicted probability distribution of each sample and its class soft label to identify noisy labels. The predicted probability distributions of clean samples tend to be closer to class soft labels than that of noisy ones.}
	\label{fig1}
\end{figure}

\section{Introduction}
\label{Introduction}
\chapteri{R}ecently, deep neural networks (DNNs) have achieved satisfying performance in various image recognition challenges\upcite{pei2022hierarchical,wang2021covid,wang2021advances,zhang2020advances,zhu2019infrared,deng2018adaptive,zhu2020tnlrs,deng2021infrared}, \eg ImageNet\upcite{imagenet} and COCO\upcite{lin2014microsoft}. Particularly, the impressive results typically rely on the availability of large-scale and well-labeled datasets. Unfortunately, high-quality and reliable annotations can be laborious and expensive\upcite{yao2021non,chen2023multi,liuhf2023}, even not always available in some domains such as fine-grained visual recognition due to the requirement of expert knowledge\upcite{tian2024unsupervised}. To address the expensive-annotation problem, a promising solution is
straightforwardly utilizing data from web or multimedia to conduct large scale datasets\upcite{yao2017exploiting,yao2019towards,yao2020exploiting,2019dynamically,xiao2015learning}. For example, WebFG-496\upcite{sun2021webly} leverages free web images as training data with keywords as labels, while YFCC100M\upcite{yfcc100m} and Youtube-8m\upcite{youtube} contain millions of media objects.
Nevertheless, annotations from web or multimedia only provide unreliable supervision and inevitably contain label noise. According to the memorization effect\upcite{arpit2017closer,zhang2016understanding}, DNNs would perfectly fit the training set with noisy labels and consequently degrade generalization.

To tackle this issue, researchers have proposed a number of methods for combating noisy labels\upcite{sun2022pnp,liu2015classification,zhang2021robust,zhang2020data,liu2021exploiting,zhang2022guided,ren2018learning,shu2019meta,sheng2023adaptive}. An active research direction is to investigate noise-robust methods which aim to reduce contributions of false-labeled samples in model optimization, \eg robust loss functions\upcite{wang2019symmetric,ma2020normalized,zhang2018generalized},  early-learning\upcite{xia2020robust}, label smoothing (LS)\upcite{szegedy2016rethinking}, and online label smoothing (OLS) regularization\upcite{zhang2021delving}. This type of method does not involve specific designs for noisy labels and therefore becomes flexible in practical application. However, since noisy labels are not explicitly coped with, noise-robust approaches still inevitably suffer from the performance drop caused by label noise.

Another intuitive research direction is to perform noise-cleaning which aims to correct or discard mislabeled samples to purify datasets. For example, label correction methods aim to revise false labels through noise transition matrix estimation\upcite{goldberger2016training} or label re-assignment\upcite{li2019dividemix}. Sample selection works\upcite{malach2017decoupling,han2018co,wei2020combating,liuhf2022,cai2023robust,cai2023co} typically select instances in manual-defined criteria, \eg the small-loss principle\upcite{gui2021towards} that regards images with small losses as clean data. Recently, some hybrid researches\upcite{song2019selfie,yao2021jo} combine label correction and sample selection methods for more efficient noise-cleaning. However, these approaches tend to be designed and tested on synthetic noisy datasets such as CIFAR\upcite{krizhevsky2009learning}, and typically do not take the real-world scenario into consideration, 
Consequently, they tend to be less practical on real-world noisy datasets.

To this end, we propose a simple yet effective approach termed GRIP (\textbf{G}roup \textbf{R}egularization and \textbf{I}nstance \textbf{P}urification) to boost instance purification via group regularization. Specifically, the proposed group regularization strategy estimates the soft label of each class to provide additional supervision. It guides the network to learn inter-class similarities and improves robustness by preventing overfitting noisy labels. Resorting to estimated class soft labels, an instance purification strategy is applied to specifically clean noisy labels from the entire dataset in a global manner. It measures the similarity between the predicted probability distribution of each sample and its estimated class soft label to identify noisy and revisable labels.
Subsequently, noisy samples are discarded and revisable instances are re-labeled by model prediction then utilized for training.
How group regularization benefits instance purification is visualized in Fig.~\ref{fig1}. Owing to the regularization strategy, clean samples and corresponding class centers are encouraged to be closer and therefore it is easier to perform noise identification.

Owing to operations from both group and instance aspects, GRIP integrates the advantages of noise-robust and noise-cleaning methods for tackling noisy labels. To sum up, our contributions are as follows:

\begin{enumerate}
	\item We propose a group regularization strategy to estimate class soft labels for benefiting instance purification. It also improves the robustness against noisy labels from the category aspect and greatly boosts the model generalization.
	
	\item We propose an instance purification strategy resorting to estimated class soft labels. It globally identifies noisy and revisable labels from the entire dataset. Experimental results demonstrate that it surpasses the widely-used small-loss principle in noise identification on real-world noisy datasets.
	
	\item Our approach integrates the advantages of noise-robust and noise-cleaning methods. Comprehensive experimental results demonstrate that GRIP significantly outperforms state-of-the-art methods on both synthetic and real-world noisy datasets.
\end{enumerate}

\begin{figure*}[t]
	\centering
	\includegraphics[width=0.98\textwidth]{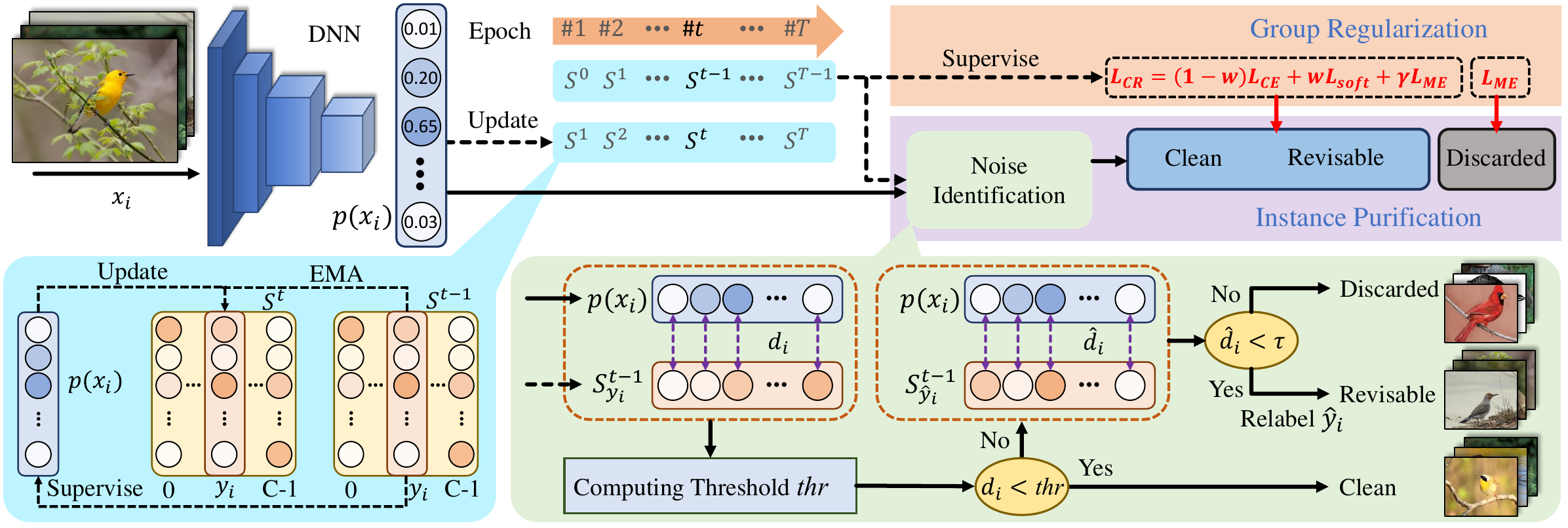}
	\caption{The framework of our proposed approach with Web-bird\upcite{sun2021webly} as an example. In each epoch $ t $, the network produces a probability $ p(x_{i}) $ for each image $ x_{i} $. Then $ p(x_{i}) $ updates the soft label of its class, and EMA is utilized to smooth the update. The estimated class soft labels $ S $ are leveraged in the noise identification and provide supervision through $ \mathcal{L}_{Soft} $. In noise identification, we compute the JS divergence $ d_{i} $ between probability $ p(x_{i}) $ and soft label $ S^{t-1}_{y_{i}} $ to select clean samples. As for noisy ones, we compute the JS divergence $ \hat{d}_{i} $ between probability $ p(x_{i}) $ and soft label of its prediction $ S^{t-1}_{\hat y_{i}} $ to divide revisable and discarded instances. The prediction $ \hat{y}_{i} $ is assigned as the pseudo label for each revisable sample. Finally, clean and revisable images are trained using $ \mathcal{L}_{GR} $. $ \mathcal{L}_{ME} $ is applied on discarded ones as regularization.}
	\label{framework}
\end{figure*}

\section{Related Works}
\subsection{Noise-Robust}
Noise-robust learning approaches directly train models using noisy labeled data. They aim to become insensitive to the presence of noisy labels\upcite{peng2020suppressing}. One typical branch is to develop robust loss functions to overcome the problem that cross-entropy loss is sensitive to samples with corrupted labels\upcite{ghosh2017robust}. For example, Wang \al proposed to leverage a noise-robust reverse cross-entropy\upcite{wang2019symmetric} to symmetrically boost cross-entropy loss. Ma \al\upcite{ma2020normalized} proposed a framework to build new loss functions by combining active and passive robust loss functions. Zhang \al\upcite{zhang2018generalized} proposed a generalization of mean absolute error and cross-entropy loss. However, since these robust loss functions typically aim to deal with noisy labels through under-learning, they inevitably underfit clean samples. Another branch is to employ regularization to improve robustness. For example, LS\upcite{szegedy2016rethinking} built soft labels by combining a one-hot and uniform distribution to provide regularization. OLS\upcite{zhang2021delving} further improved LS by replacing the uniform distribution with a more reasonable probability on non-target categories. However, these approaches do not explicitly tackle noisy labels, leading to a suboptimal performance.

\subsection{Noise-Cleaning}

Noise-cleaning methods aim to tackle noisy labels by discarding or relabeling them. An intuitive type of research is label correction that corrects noisy labels. For example, several works\upcite{goldberger2016training} tried to correct noisy labels by estimating the noise transition matrix. However, it is difficult to estimate the accurate noise transition matrix. Furthermore, these label correction methods are unable to deal with out-of-distribution (OOD) instances\upcite{zhang2020data} whose true labels do not belong to the training set. This drawback restricts their application in the real-world scenario. Another typical idea of combating noisy labels is to perform sample selection that identifies and removes corrupted data through proper criteria\upcite{cai2023robust,cai2023co,sun2020crssc}. For example, Co-teaching\upcite{han2018co} utilized the small-loss principle and let peer networks select noisy samples for each other. JoCoR\upcite{wei2020combating} claimed that different models tend to disagree on false labeled samples and leveraged this principle for noise identification. Nevertheless, the above approaches typically performed sample selection within each mini-batch with a fixed drop rate. Jo-SRC\upcite{yao2021jo} and Zhang \al\upcite{AAAI2020} claimed that noise ratios in different mini-batches inevitably fluctuate in the training process. To overcome this problem, they replaced the widely-used mini-batch noise identifying strategy with a global one for the purpose of stabilizing the selection results. A growing number of methods such as SELFIE\upcite{song2019selfie} and Co-learning\upcite{tan2021co} combined label correction and sample selection to further boost the performance. Some state-of-the-art noise-cleaning approaches also adopted noise-robust techniques. For example, Jo-SRC\upcite{yao2021jo} and DivideMix\upcite{li2019dividemix} leveraged LS trick and mix-up\upcite{berthelot2019mixmatch}, respectively.

\section{The Proposed Method}
\subsection{Preliminary}
Generally, we train a DNN on a noisy dataset $ \mathcal{D}=\{(x_{i},y_{i})| 1 \leq i \leq N \} $ with $ C $ classes, where $ x_{i} $ and $ y_{i} $ denote the $ i $-th training sample and the corresponding label, respectively. We define the label distribution $ q $ of the one-hot label $ y_{i} $ as $ q(c =y_{i}|x_{i}) = 1 $ and $ q(c \neq y_{i}|x_{i}) = 0 $.
The DNN model takes each sample $ x_{i} $ as input and predicts a probability $ p(c|x_{i}) $ for each class $ c $ using the softmax function. The cross-entropy training loss between the predicted probability distributions $ p $ of training images and their corresponding label $ q $ is written as
\begin{equation}
	\mathcal{L}_{CE}=- \frac{1}{N} \sum_{i=1}^{N} \sum_{c=1}^{C} q(c|x_{i}) \log{p(c|x_{i})}.
\end{equation}
Since commonly-used cross-entropy loss is proved to be sensitive to noisy labels\upcite{ghosh2017robust} and DNNs can perfectly memorize noisy samples\upcite{arpit2017closer,zhang2016understanding}, deep models tend to suffer from label noise when trained using the noisy dataset $ \mathcal{D} $ with $ \mathcal{L}_{CE} $. 

\begin{figure}[t]
	\centering
	\includegraphics[width=0.49\textwidth]{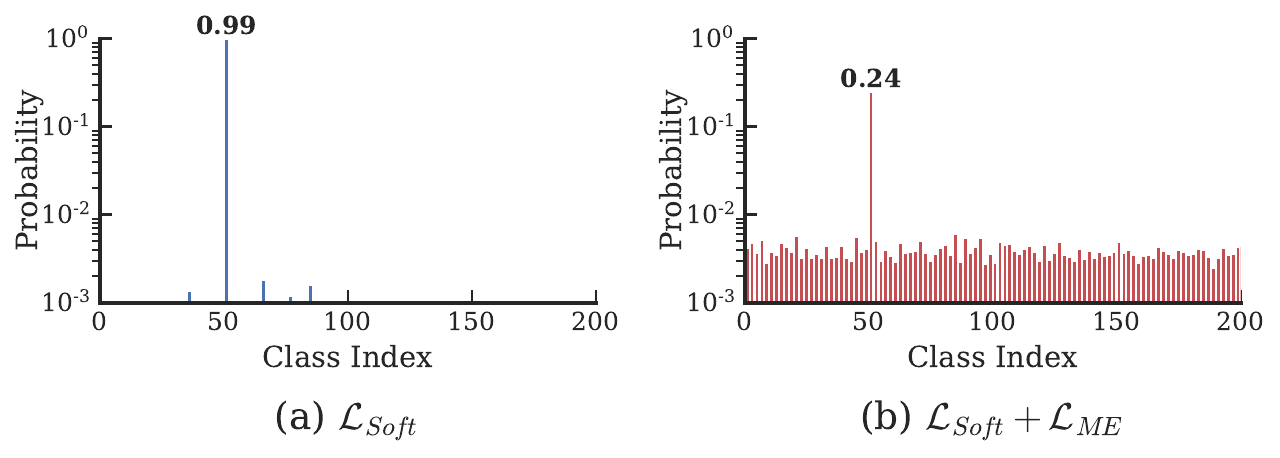}
	\caption{Label distributions of $ \mathcal{L}_{Soft} $ (a) and $ \mathcal{L}_{Soft} + \mathcal{L}_{ME} $ (b) on Web-bird. We scale the $ y $-axis using the log function for visualization. Soft labels are generated during the training process of a ResNet-18 model.}
	\label{fig3}
\end{figure}

\subsection{Group Regularization}
Our learning framework is illustrated in Fig.~\ref{framework}. Motivated by OLS\upcite{zhang2021delving}, we adopt a noise-robust strategy through regularizing the training process at the category level. We define $ S = \{ S^{0}, S^{1},\cdots, S^{t},\cdots, S^{T-1}\} $ as the collection of soft labels of each class for $ T $ training epochs. For each epoch $ t $, $ S^{t} $ is a $ C \times C $ matrix, whose columns correspond to soft labels and are initialized to zero. Given an input image $ x_{i} $, if the classification is in accord with its label $ y_{i} $ , the soft label $ S_{y_{i}}^{t} $ of the target class $ y_{i} $ will be updated using the predicted probability $ p(x_{i}) $ through
\begin{equation}
	S_{y_{i},c}^{t} = \frac{1}{M} \sum_{j=1}^{M} p(c|x_{j}),
	\label{eq2}
\end{equation}
where $ c \in \{1,2,\cdots, C \} $ denotes the indexes of $ S_{y_{i}}^{t} $ and $ M $ indicates the number of correctly predicted samples with label $ y_{i} $. According to Eq. (\ref{eq2}), each soft label is the average predicted probability of correctly classified samples belonging to one class.

To stabilize the estimated class soft labels $ S^{t} $, we further apply an exponential moving average (EMA) strategy through
\begin{equation}
	\begin{aligned}
		S_{y_{i},c}^{t} = m S_{y_{i},c}^{t-1} + \frac{1-m}{M} \sum_{j=1}^{M} p(c|x_{j}),
	\end{aligned}
	\label{eq3}
\end{equation}
where $ m $ denotes the momentum that controls the weight on the previous result. EMA smoothes out $ S^{t} $ by alleviating the issue that probabilities predicted by the model can be unstable in the training process.
After obtaining the soft label, we utilize $ S^{t-1} $ to supervise the training process in each epoch $ t $. Then the soft training loss can be formulated as
\begin{equation}
	\mathcal{L}_{Soft} =- \frac{1}{N} \sum_{i=1}^{N} \sum_{c=1}^{C} S_{y_{i},c}^{t-1} \log{p(c|x_{i})}.
\end{equation}
Similar to LS, $ \mathcal{L}_{Soft} $ assigns weights to non-target categories. Consequently, it reduces overfitting and improves the robustness against noisy labels. Moreover, it encourages DNNS to learn inter-class similarities and benefits challenging image recognition tasks, \eg fine-grained classification\upcite{zhang2021delving}.

However, we find that $ S_{y_{i}}^{t} $ tends to approach the one-hot label distribution where $ y_{i} $ has a large value while other categories only share tiny weights as shown in Fig.~\ref{fig3} (a). This behavior may derive from the strong fitting ability of cross-entropy loss. To address this issue, we utilize the maximum entropy principle\upcite{dubey2018maximum} as a strong regularization. It forces the model prediction to be less confident. The maximum entropy loss can be formulated as
\begin{equation}
	\begin{aligned}
		\mathcal{L}_{ME} &= -\frac{1}{N} \sum_{i=1}^{N} \mathrm{H}(p(x_{i}))\\
		&= \frac{1}{N} \sum_{i=1}^{N} \sum_{c=1}^{C}p(c|x_{i})\log{p(c|x_{i})}.
	\end{aligned}
	\label{eq4}
\end{equation}

Since $ \mathcal{L}_{ME} $ aims to increase the entropy of prediction $ p(x_{i}) $, it leads to a more reasonable soft label as illustrated in Fig.~\ref{fig3}~(b). More importantly, since $ \mathcal{L}_{ME} $ makes predictions less confident, it further reduces the risk of overfitting noisy labels and guides DNNs to become more noise-robust.

Finally, both the hard and soft labels are leveraged as supervision with the maximum-entropy principle applied as regularization. The total training loss of our group regularization strategy can be represented by
\begin{equation}
	\mathcal{L}_{GR} = (1-w) \mathcal{L}_{CE} + w \mathcal{L}_{Soft} + \gamma \mathcal{L}_{ME},
	\label{eq6}
\end{equation}
where $ w $ and $ \gamma $ are weights to balance contributions of $ \mathcal{L}_{CE} $, $ \mathcal{L}_{Soft} $, and $ \mathcal{L}_{ME} $. Owing to $ \mathcal{L}_{Soft} $ and $ \mathcal{L}_{ME} $, our group regularization strategy guides DNNs to become less sensitive to noisy labels and boosts their robustness. Furthermore, class soft labels generated by group regularization strategy can be leveraged for instance purification.

\subsection{Instance Purification}

After obtaining class soft labels, we can perform data purification from the instance aspect.
According to the memorization effect\upcite{arpit2017closer,zhang2016understanding}, DNNs learn the clean and easy pattern in the initial epochs. Accordingly, clean instances tend to have more contributions to the estimation of class soft labels at the early stage. Moreover, our group regularization strategy reduces the contribution of noisy labels in model optimization by significantly improving noise-robustness.
Therefore, generated class soft labels should be closer to predictions of clean samples than that of noisy ones. Based on this phenomenon, we can build a noise identification criterion.

Inspired by Jo-SRC\upcite{yao2021jo}, we adopt the Jensen-Shannon (JS) divergence between each probability $ p(x_{i}) $ and its class soft label $ S_{y_{i}}^{t-1} $ as our noise identification criterion through
\begin{equation}
	\begin{aligned}
		d_{i} =& \mathrm{D}_{JS}(p(x_{i})||S_{y_{i}}^{t-1})\\
		=& \frac{1}{2}\mathrm{D}_{K\!L}(p_{i} || \frac{p_{i} \!+\! S_{y_{i}}^{t-1}}{2}) \!+\!  \frac{1}{2}\mathrm{D}_{K\!L}(S_{y_{i}}^{t-1} || \frac{p_{i} \!+\! S_{y_{i}}^{t-1}}{2}),
	\end{aligned}
	\label{eq7}
\end{equation}
where $ D_{KL} $ is the Kullback-Leibler (KL) divergence function and $ p_{i} $ denotes the simplified form of $ p(x_{i}) $. In this formula, $ d_{i} $ is a measure of the difference between predicted probability $ p_{i} $ and corresponding class soft label $ S_{y_{i}}^{t} $, where a larger value indicates a more significant difference. Moreover, $ d_{i} $ is a symmetric measure and bounded in $ [0, 1] $ when using a base $ 2 $ logarithm\upcite{lin1991divergence}. 

\begin{figure}[t]
	\includegraphics[width=0.48\textwidth]{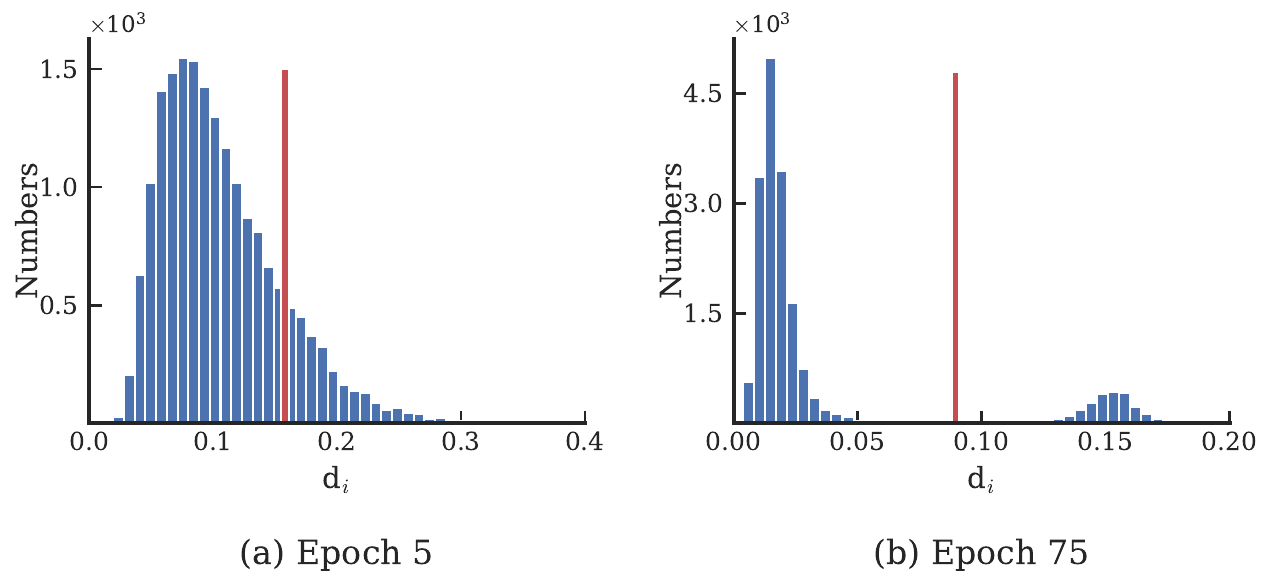}
	\caption{The distribution of $ d $ in epoch $ 5 $ (a) and $ 75 $ (b) during the training process of a ResNet-18 model on Web-bird. The red bar indicates the threshold $ thr $. As the training proceeds, $ d_{i} $ decreases and $ thr $ automatically adapts to the change of $ d_{i} $. The distribution becomes more discrete because noisy samples are discarded.}
	\label{fig4}
\end{figure}

Since clean samples tend to be closer to their class soft labels, they should have smaller values of $ d_{i} $ than noisy instances. Accordingly, we can utilize a threshold to separate clean instances from noisy ones. We define the threshold $ thr $ for each epoch $ t $ as
\begin{equation}
	thr = \textbf{mean}(d) + \alpha \cdot \textbf{std}(d),
	\label{eq8}
\end{equation}
where $ d=\{d_{1},d_{1}, \cdots ,d_{i}, \cdots,d_{N}\} $ indicates the collection of $ d_{i} $ for the entire dataset, $ \textbf{mean}(\cdot) $ and $ \textbf{std}(\cdot) $ denote computing average value and standard deviation respectively, and $ \alpha $ is a hyperparameter. Specifically, for each training batch $ \mathcal{B} $, we divide it into a clean set $ \mathcal{B}_{clean} $ and a noisy set $ \mathcal{B}_{noisy} $ after a warm-up period $ t_{m} $ by

\begin{equation}
	\begin{aligned}
		\mathcal{B}_{clean} &=\{(x_{i},y_{i})| d_{i} \leq thr, t \geq t_{m} \}, \\
		\mathcal{B}_{noisy} &=\{(x_{i},y_{i})| d_{i} > thr, t \geq t_{m} \}.
	\end{aligned}
	\label{eq9}
\end{equation}

Once noisy samples are identified, we further perform label re-assignment on the noisy samples $ \mathcal{B}_{noisy} $ to leverage revisable ones. Specifically, we first calculate the JS divergence between the probability $ p(x_{i}) $ of each instance in $ \mathcal{B}_{noisy} $ and soft label $ S_{\hat{y}_{i}}^{t-1} $ of its predicted class $ \hat{y}_{i} $ through 
\begin{equation}
	\hat{d}_{i} = \mathrm{D}_{JS}(p(x_{i})||S_{\hat{y}_{i}}^{t-1}), x_{i} \in \mathcal{B}_{noisy}.
\end{equation}
Then, since $ \hat{d}_{i} $ is bounded in $ [0, 1] $, we can utilize a hard threshold $ \tau $ to select revisable samples in $ \mathcal{B}_{noisy} $ through
\begin{equation}
	\begin{aligned}
		\mathcal{B}_{relabel} &= \{(x_{i}, \hat{y}_{i})| x_{i} \in \mathcal{B}_{noisy}, \hat{d}_{i} < \tau \}, \\
		\mathcal{B}_{discard} &= \{(x_{i}, y_{i})| x_{i} \in \mathcal{B}_{noisy}, \hat{d}_{i} \geq \tau \}.
	\end{aligned}
	\label{eq11}
\end{equation}
Eq.~(\ref{eq11}) indicates that we regard a noisy sample as a revisable one when its probability $ p(x_{i}) $ is close enough to the soft label of its predicted category $ S_{\hat{y}_{i}}^{t-1} $. Its predicted class $ \hat{y}_{i} $ is assigned as pseudo label.

Finally, we jointly utilize clean and relabeled samples for training by Eq.~(\ref{eq6}). Furthermore, we apply $ \mathcal{L}_{ME} $ in Eq.~(\ref{eq4}) on discarded images to encourage the model to generate even probabilities for them. This design guides predicted probabilities of noisy samples to become more different from estimated class soft labels and therefore improves the reliability of noise-cleaning in the following epochs.
Details of our proposed GRIP are demonstrated in Algorithm \ref{algorithm}.

\subsection{Discussion}

\subsubsection{Comparison with OLS}
Our group regularization strategy is motivated by OLS\upcite{zhang2021delving}. Compared with OLS, our approach applies maximum entropy loss as regularization to smooth the predicted probability. It addresses the problem that the estimated soft label tends to approach the one-hot distribution. Furthermore, EMA utilizes previous results to stabilize the process of soft label estimation. Consequently, our group regularization strategy becomes more practical and efficient than OLS.

\subsubsection{Dynamic and Fixed Thresholds}
Notably, in the initial stages of training, a significant disparity between the predicted distribution and soft labels is observed, yielding generally large values of d. Conversely, as training progresses, there is a trend towards smaller values of d. The use of a fixed threshold would result in the model initially selecting too few clean samples, leading to a low recall rate. Furthermore, in later stages, there arises the issue of selecting too many samples, encompassing a substantial amount of noisy false positives and resulting in a low precision rate. To address these challenges, we employ the mean and standard deviation of the distance to formulate the threshold for sample selection criteria, thereby enhancing the flexibility and adaptability of our approach. This selection strategy, widely embraced by numerous classical algorithms\upcite{patel2023adaptive}, attests to its effectiveness and generalizability.

As shown in Fig.~\ref{fig4} (a) and (b), $ d_{i} $ decreases as the DNN gradually fits the dataset in the training process. Accordingly, we utilize a dynamic threshold $ thr $, whose value depends on the distribution of $ d $. The mean value, which roughly separates "low" and "high", can be utilized as a rough threshold to identify clean and noisy labels. To further adjust the threshold, we leverage the standard deviation controlled by hyperparameter $ \alpha $ as the offset to the mean. Compared with a fixed offset, the standard deviation can automatically adapt to the distribution of $ d $. It will be small if the data distribution is dense and become large otherwise. Briefly, the threshold is jointly controlled by mean as well as standard deviation. It can adapt to the changing $ d_{i} $ and select samples with lower $ d_{i} $ for training. Similar to $ d_{i} $, $ \hat{d}_{i} $ also gradually decreases in the training process. Contrary to threshold $ thr $, we fix $ \tau $ for the label re-assignment. It guides the number of relabeled samples to grow from a small value as the training proceeds. 
As the model becomes more robust, more noisy samples can be relabeled for training. This progressive re-assignment strategy controls the number of relabeled samples and stabilizes the training process.

\subsubsection{Selection Criterion} Since $ thr $ is obtained according to the distribution of $ d $ on the entire dataset, the selection operation is performed in a global view. This global selection strategy can alleviate the problem that noise ratios in different mini-batches inevitably fluctuate\upcite{yao2021jo,AAAI2020}. A mini-batch is a sampling of the entire dataset. If the batch size is large enough (\eg >100), the noise ratio fluctuation problem may not be severe. However, on the contrary, if the batch size has to be small due to the large size of the model or input image, this problem can harm the noise-cleaning performance. We will illustrate the advantage of the global selection strategy over a mini-batch one in experiments.

Compared with the popular small-loss principle, our selection criterion takes the advantage of soft labels in noise identification. The soft label encodes more information than training losses and therefore tends to be more reliable in selecting samples.

\begin{algorithm}[t]
	\SetAlgoLined
	\caption{Group Benefits Instance for Data Purification}
	\KwInput{Network $ \theta $, web set $ \mathcal{D} $, warm-up epoch $ t_{m} $, momentum $ m $, weight $  w $ and $ \gamma $, hyperparameter $ \alpha $, threshold $ \tau $ and maximum epochs $ T $.}
	\textbf{Initialize} Network $ \theta $.\\
	\For{$t = 1, 2,..., T$}{
		\For {each mini-batch $ \mathcal{B} $ in $ \mathcal{D} $}{
			\uIf {$ t < t_{m} $}
			{
				Compute $ \mathcal{L} $ using $ \mathcal{B} $ by Eq.~(\ref{eq6}).
			}
			\Else
			{
				Update $ d_{i} $ according to Eq.~(\ref{eq7}).\\
				Update $ thr $ according to Eq.~(\ref{eq8}).\\
				Obtain $ \mathcal{B}_{clean} $ and $ \mathcal{B}_{noisy} $ by Eq.~(\ref{eq9}).\\
				Obtain $ \mathcal{B}_{relabel} $ and $ \mathcal{B}_{discard} $ by Eq.~(\ref{eq11}).\\
				Compute $ \mathcal{L}_{1} $ using $ \mathcal{B}_{clean} $ and $ \mathcal{B}_{relabel} $  according to Eq.~(\ref{eq6}).\\
				Compute $ \mathcal{L}_{2} $ using $ \mathcal{B}_{discard} $ by Eq.~(\ref{eq4}).\\
				Sum $ \mathcal{L} = \mathcal{L}_{1} + \mathcal{L}_{2} $.
			}
			Update $ S^{t} $ according to Eq.~(\ref{eq3}).\\
			Update $ \theta = \mathrm{SGD}(\mathcal{L};\theta) $.
		}
	}
	\KwOutput{Updated network $ \theta $.}
	\label{algorithm}
\end{algorithm}


\section{Experiments on Synthetic Noisy Datasets}

In order to demonstrate the superiority and applicability of our approach, we conduct experiments on both synthetic and real-world datasets.
In this experiment, we first evaluate our approach on synthetic noisy datasets for coarse-grained classification to validate its robustness under different types of noisy labels and varying noise ratios.

\begin{table*}[t]
	\centering
	\caption{Average Classification Accuracy (\textbf{ACA $\%$}) on CIFAR-10 over the last 10 epochs.}
		\setlength{\tabcolsep}{2pt}
		\scalebox{0.95}{
			\begin{tabular}{c|c|c|c|c|c|c|c}
				\toprule
				Noise Ratio $ \epsilon $ & Backbone & Decoupling\upcite{malach2017decoupling}& Co-teaching\upcite{han2018co}& Co-teaching+\upcite{yu2019does} & JoCoR\upcite{wei2020combating} & SELC\upcite{selc} & \textbf{GRIP} \\
				\midrule
				$\text{Symmetry} -20\%$  & 69.18 $\pm$ 0.52 & 69.32 $\pm$ 0.40 & 78.23 $\pm$ 0.27 & 78.71 $\pm$ 0.34  & 85.73 $\pm$ 0.19  & 87.08 $\pm$ 0.07  & \textbf{88.83} $\pm$ 0.09 \\
				$\text{Symmetry} -50\%$  & 42.71 $\pm$ 0.42 & 40.22 $\pm$ 0.30 & 71.30 $\pm$ 0.13 & 57.05  $\pm$ 0.54  & 79.41 $\pm$ 0.25  & 81.66 $\pm$ 0.11   & \textbf{84.59} $\pm$ 0.07  \\
				$\text{Symmetry} -80\%$&28.67 $\pm$ 0.47 &26.76 $\pm$ 0.35 & 34.53 $\pm$ 0.28 & 29.05 $\pm$ 0.21  & 47.74 $\pm$ 0.25 & 54.58 $\pm$ 0.15 & \textbf{59.32} $\pm$ 0.07 \\
				$\text{Asymmetric}-40\%$  & 69.43 $\pm$ 0.33 & 68.72 $\pm$ 0.30 & 73.78 $\pm$ 0.22 & 68.84 $\pm$ 0.20  & 76.36  $\pm$ 0.49  & 78.90 $\pm$ 0.12  & \textbf{80.82} $\pm$ 0.28 \\
				\bottomrule
		\end{tabular}}
		\label{cifar10}
\end{table*}

	\begin{table*}[t]
	\centering
	\caption{Average Classification Accuracy (\textbf{ACA $\%$}) on CIFAR-100 over the last 10 epochs.}
	\setlength{\tabcolsep}{2pt}
	\scalebox{0.95}{
		\begin{tabular}{c|c|c|c|c|c|c|c}
			\toprule
			Noise Ratio $ \epsilon $  & Backbone & Decoupling\upcite{malach2017decoupling}& Co-teaching\upcite{han2018co}& Co-teaching+\upcite{yu2019does} & JoCoR\upcite{wei2020combating} & SELC\upcite{selc} & \textbf{GRIP} \\
			\midrule
			$\text{Symmetry} -20\%$  & 35.14 $\pm$ 0.44 & 33.10 $\pm$ 0.12 & 43.73 $\pm$ 0.16 & 49.27 $\pm$ 0.03  & 53.01 $\pm$ 0.04  & 55.44 $\pm$ 0.09  & \textbf{61.40} $\pm$ 0.13 \\
			$\text{Symmetry} -50\%$  & 16.97 $\pm$ 0.40 & 15.25 $\pm$ 0.20 & 34.96 $\pm$ 0.50 & 40.04 $\pm$ 0.70  & 43.49 $\pm$ 0.46  & 46.73 $\pm$ 0.08  & \textbf{51.27} $\pm$ 0.15  \\
			$\text{Symmetry} -80\%$  &  9.67 $\pm$ 0.34 &  8.69 $\pm$ 0.09 & 23.62 $\pm$ 0.38 & 27.08 $\pm$ 0.51  & 31.92 $\pm$ 0.27  & 34.51 $\pm$ 0.15  & \textbf{39.67} $\pm$ 0.15  \\
			$\text{Asymmetric}-40\%$  & 27.29 $\pm$ 0.25 & 26.11 $\pm$ 0.39 & 28.35 $\pm$ 0.25 & 33.62 $\pm$ 0.39  & 32.70 $\pm$ 0.35 & 45.19 $\pm$ 0.12  & \textbf{53.48} $\pm$ 0.11 \\
			\bottomrule
	\end{tabular}}
	\label{cifar100}
\end{table*}
	
	\subsection{Datasets and Evaluation Metric}
	\label{cifar_setting}
	
	We follow the common experimental settings as in recent works\upcite{han2018co,wei2020combating} and manually corrupt CIFAR-10 and CIFAR-100\upcite{krizhevsky2009learning} to create synthetic noisy datasets. Specifically, we adopt symmetric and asymmetric noisy labels with noise transition matrices with a transition matrix $ Q $. 
	
	We generate symmetric noisy labels by uniformly flipping a given ratio of training labels to other classes. Asymmetric label noise simulates the fine-grained recognition task with noisy labels, where very similar classes may confuse annotators. In this setting, we only flip a specific set of classes, \eg flipping CAT to DOG in  CIFAR-10. The definition of transition matrix $ Q $ is as follows:
	
	\begin{equation}
		\begin{aligned}
			Q_{symmetry}&=
			\begin{pmatrix}
				1 - \epsilon 		 & \frac{\epsilon}{n-1} & \dots  & \frac{\epsilon}{n-1} & \frac{\epsilon}{n-1}\\ 
				\frac{\epsilon}{n-1} & 1 - \epsilon  		& \dots  & \frac{\epsilon}{n-1} & \frac{\epsilon}{n-1}\\
				\vdots 		 		 &   					& \ddots &   					& \vdots			  \\
				\frac{\epsilon}{n-1} & \frac{\epsilon}{n-1} & \dots  & 1 - \epsilon 	    & \frac{\epsilon}{n-1}\\
				\frac{\epsilon}{n-1} & \frac{\epsilon}{n-1} & \dots  & \frac{\epsilon}{n-1} & 1 - \epsilon		  \\
			\end{pmatrix},\\
			Q_{asymmetry}&=
			\begin{pmatrix}
				1 - \epsilon 		 & \epsilon 			& 0        & \dots			    & 0			\\ 
				0 					 & 1 - \epsilon  		& \epsilon & \dots				& 0			\\
				\vdots 		 		 &   					& \ddots   & \ddots				& \vdots			  \\
				0					 & 0					& \dots    & 1 - \epsilon 	    & \epsilon  \\
				\epsilon			 & 0 				    & \dots    & 0				    & 1 - \epsilon	  \\
			\end{pmatrix},
		\end{aligned}
		\nonumber
	\end{equation}
	where $ n $ and $ \epsilon $ denote the number of the class and noise ratio, respectively.
	For evaluating the model classification performance, we take Average Classification Accuracy (\textbf{ACA}) as the evaluation metric.
	
	\subsection{Implementation Details} 
	
	Following JoCoR\upcite{wei2020combating}, we adopt a 7-layer CNN network architecture and leverage Adam optimizer with momentum set to $ 0.9 $ for training. We train the network for $ 200 $ epochs with batch size set to $ 128 $. The learning rate is initialized to $ 0.001 $ and starts to linearly decrease to $ 0 $ after $ 80 $ epochs.
	As for hyper-parameters, we set momentum $ m $ for EMA and weight $ w $, $ \gamma $ to $ 0.5 $, $ 0.5 $, $ 1 $, respectively. Warm-up epoch $ t_{m} $ and hyper-parameter $ \tau $ are empirically set to $ 10 $ and $ 0.03 $, respectively. As for $ \alpha $, we adopt a linear decrease from $ 1 $ to $ 0.2 $ in $ 5 $ epochs to smooth the discarding process, which is similar to the increasing drop rate trick in Co-teaching\upcite{han2018co}. All experiments are conducted on one NVIDIA Tesla V100 GPU.

	\subsection{Baseline Methods}
	
	To evaluate our approach on synthetic datasets, we compare our approach with the following state-of-the-art algorithms: Decoupling\upcite{malach2017decoupling}, Co-teaching\upcite{han2018co}, Co-teaching+\upcite{yu2019does}, JoCoR\upcite{wei2020combating}, SELC\upcite{selc} and the backbone network straightforwardly trained on noisy datasets. As the work of JoCoR\upcite{wei2020combating} has reproduced most other baselines\upcite{malach2017decoupling,han2018co,yu2019does}, we directly copy results from it for simplicity. For SELC\upcite{selc}, we re-execute their code and report the experimental results. Our method is trained using the same experimental settings as the above methods, and therefore the comparisons are fair.

	\begin{figure*}[t]
		\includegraphics[width=\linewidth]{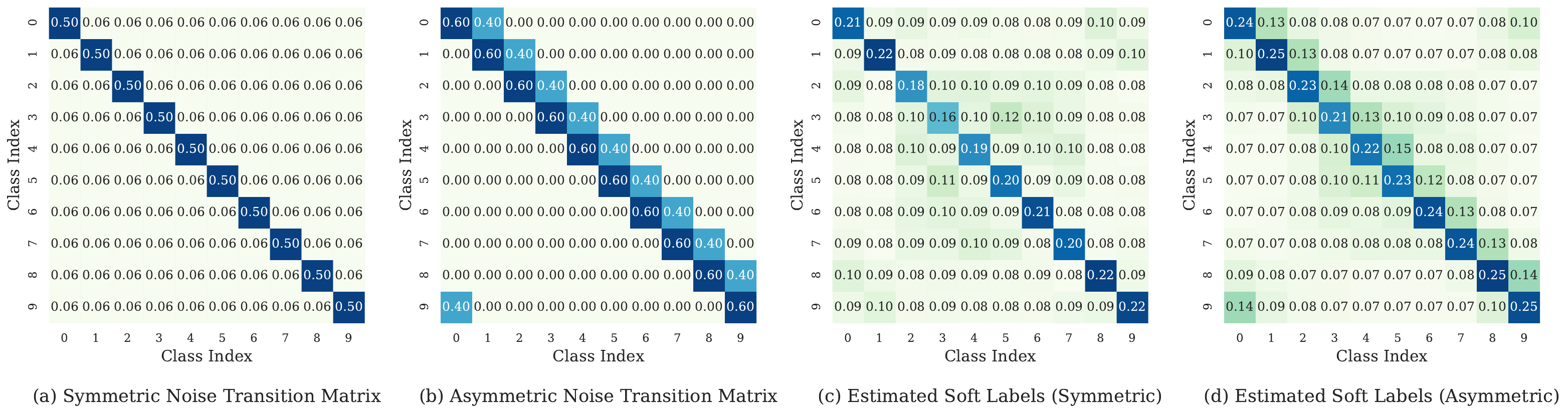}
		\caption{The symmetric (a) and asymmetric (b) noise transition matrices and corresponding estimated soft labels after the warm-up period on CIFAR-10 ((c) and (d)). The noise ratios $ \epsilon $ are set to $ 0.5 $ and $ 0.4 $ for symmetric and asymmetric noise, respectively.}
		\label{matrix_cifar}
	\end{figure*}
	
	\subsection{Experimental Results and Analysis}
	
	\subsubsection{Results on CIFAR-10} We demonstrate the test accuracy of each method on CIFAR-10 in Table~\ref{cifar10}. As we can see from Table~\ref{cifar10}, GRIP consistently outperforms other approaches in all three noisy settings. Specifically, GRIP surpasses the best results of baselines by $ 1.75 \% $, $ 2.93 \% $, $ 4.74 \% $ and $ 1.92 \% $ in Symmetry-$20\%$, Symmetry-$50\%$, Symmetry-$80\%$, and Asymmetry-$40\%$ cases, respectively. From experimental results, we can conclude that our approach is effective on synthetic noisy datasets for simple coarse-grained classification tasks.
	
	\subsubsection{Results on CIFAR-100} The test accuracy of each approach on CIFAR-100 is shown in Table~\ref{cifar100}. By observing Table~\ref{cifar100}, we can find that GRIP achieves the highest test accuracies among all methods. Specifically, the improvements over JoCoR become $ 5.96 \% $, $ 4.54 \% $, $ 5.16 \% $ and $ 8.29 \% $ in Symmetry-$20\%$, Symmetry-$50\%$, Symmetry-$80\%$, and Asymmetry-$40\%$ cases, respectively. Our great superiority in the Asymmetry-$40\%$ case deserves attention. Since asymmetric noise is a simulation of the real-world label noise\upcite{li2019dividemix}, our advantage in Asymmetry-$40\%$ cases indicates that GRIP is promising on real-world noisy datasets. 
	
	Moreover, by comparing the results on CIFAR-10 and CIFAR-100, we can find that GRIP shows more significant improvements over baselines as the classification task becomes more challenging (from 10 to 100 classes). This experimental result demonstrates the superiority of our method in more complicated classification tasks with noisy labels.
	

	\subsection{Noise Matrix and Soft Labels}
	
	In this subsection, we visualize the noise transition matrix $ Q $ and soft labels $ S $ estimated by GRIP to illustrate the noise-robustness of our approach. To be detailed, we investigate the Symmetry-$50\%$ and Asymmetry-$40\%$ cases and demonstrate soft labels estimated after the warm-up period in Fig.~\ref{matrix_cifar}.
	
	From Fig.~\ref{matrix_cifar}, we can observe that in the Symmetry-$50\%$ case, all non-target classes of soft labels share similar weights (Fig.~\ref{matrix_cifar} (a) and (c)), and target categories have lower weights in soft labels than in noise transition matrices. This phenomenon indicates that our approach becomes less confident on the noisy datasets and shows robustness against label noise.
	
	While in the Asymmetry-$40\%$ case, flipped noisy classes have much lower weights in soft labels than in noise transition matrices (Fig.~\ref{matrix_cifar} (b) and (d)), which indicates that our approach is less likely to overfit noisy labels. The great noise-robustness results from the maximum entropy regularization, which guides the model to become less confident on potential noisy labels. 
		\begin{table}[tb]
		\centering
		\caption{Numbers of class, training and test images, as well as estimated label accuracies of three sub-datasets.}
		\setlength{\tabcolsep}{5pt}
		\scalebox{0.97}{
			\begin{tabular}
				{r | c c c}
				\toprule
				\textbf{Dataset} & \textbf{Web-bird} & \textbf{Web-aircraft} & \textbf{Web-car}  \\
				\midrule
				Class Number     & 200    & 100   & 196   \\
				Training Images  	 & 18388  & 13503 & 21448     \\
				Test Images 	 	 & 5794   & 3333  & 8041    \\
				Label Accuracy   & 65\%   & 73\%  & 67\% \\
				\bottomrule
		\end{tabular}}
		\label{WebFG-496}
	\end{table}

		\begin{figure*}[t]
	\includegraphics[width=1.0\textwidth]{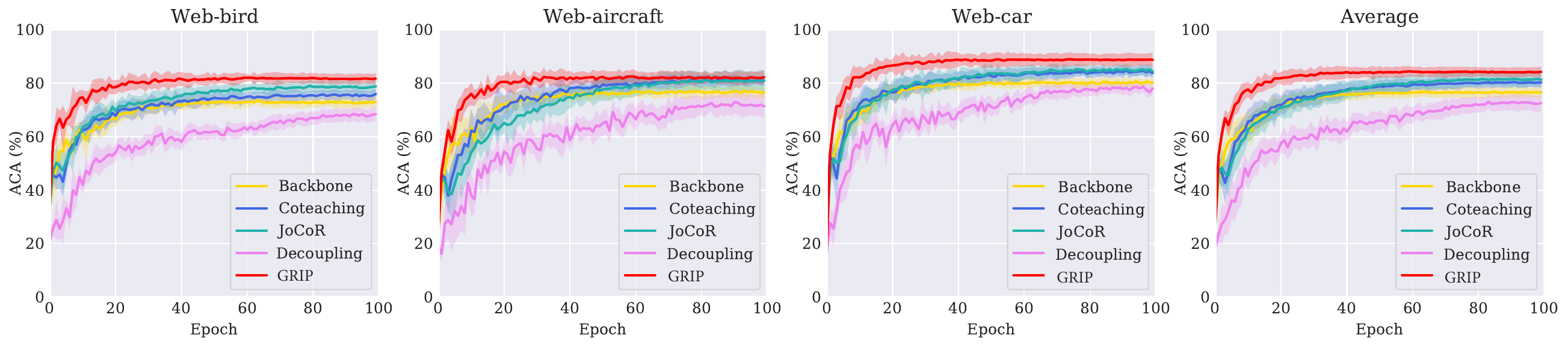}
	\caption{ACA (\%) performances of our approach and baselines on Web-bird, Web-aircraft, Web-car, and average performances.}
	\label{aca}
\end{figure*}
	\begin{table*}[t]
		\centering
		\renewcommand{\arraystretch}{1}
		\caption{ACA (\%) performances of baseline methods and ours on Web-aircraft, Web-bird, and Web-car.}
		\setlength{\tabcolsep}{8pt}
		\scalebox{0.98}{
			\begin{tabular}{r|c|c|c|c|c|c}
				\hline
				\multirow{2}{*}{\textbf{Method\:\:\:\:\:\:\:\:}} & \multirow{2}{*}{\textbf{Publication}} & \multirow{2}{*}{\textbf{Noise-cleaning}} & \multirow{2}{*}{\textbf{Noise-robust}} &  \multicolumn{3}{c}{\textbf{Datasets}} \\
				\cline{5-7}    &  &      &    & \ Web-bird \ & Web-aircraft & \ Web-car \ \\		
				\hline
				Backbone	            	  			  & -			 &  &  & 73.30  & 78.71 & 80.86\\	
				Decoupling\upcite{malach2017decoupling}    & NeurIPS 2017 & \checkmark & & 68.81   & 73.21 & 78.71\\
				Co-teaching\upcite{han2018co}              & NeurIPS 2018 & \checkmark & & 75.51   & 79.51 & 83.42\\
				Co-teaching+\upcite{yu2019does}            & ICML  2019   & \checkmark & & 70.12   & 74.80 & 76.77\\
				Sub-center\upcite{deng2020sub}             & ECCV 2020    &  & \checkmark & 75.77   & 79.80  & 82.59\\
				Self-Adaptive\upcite{huang2020self}		  &	NeurIPS 2020 & \checkmark  & & 78.49   & 77.92 & 78.19 \\
				JoCoR\upcite{wei2020combating}		  	  &	CVPR 2020    & \checkmark  & & 79.19   & 80.11 & 85.10 \\
				DivideMix\upcite{li2019dividemix}		  &	ICLR 2020    & \checkmark  & \checkmark & 74.40   & 82.48 & 84.27 \\
				PLC\upcite{zhang2021learning}   			  & ICLR 2021    & \checkmark  & & 76.22   & 79.24 & 81.87 \\
				Peer-learning\upcite{sun2021webly}	  &	ICCV 2021    &\checkmark &  & 75.37 & 78.64 & 82.48 \\
				OLS\upcite{zhang2021delving}   			  & TIP 2021     &   & \checkmark & 79.11   & 81.27 & 86.58 \\
				Jo-SRC\upcite{yao2021jo}   		  		  & CVPR 2021    & \checkmark  & & 81.52   & 82.73 & 88.13 \\
				SELC\upcite{selc}						 & IJCAI 2022    &   & \checkmark & 77.25   & 80.26 & 80.89 \\
				Co-LDL\upcite{coldl} &   TMM 2022  &  \checkmark  &  &  80.11 &  81.97 &  86.95 \\
				AGCE\upcite{asymloss} &  TPAMI 2023 &   & \checkmark &  75.54 &  82.21 &  82.76 \\
				CMW-Net-SL\upcite{cmwnet}				& TPAMI 2023    &    & \checkmark  & 77.41   & 76.48 & 79.70 \\
				\textbf{GRIP(Ours)}            	      & -            & \checkmark  & \checkmark & \textbf{82.53}   &\textbf{83.29} &\textbf{89.49}
				\\
				\hline
		\end{tabular}}
		\label{tab1}
	\end{table*}

	\section{Experiments on Real-world Noisy Datasets}
	
	In this experiment, we further evaluate our approach on more challenging web noisy datasets to demonstrate its applicability under real-world scenario.

	\subsection{Datasets and Evaluation Metric}
	
	We evaluate our approach on WebFG-496\upcite{sun2021webly}, which is designed for research on webly supervised fine-grained classification tasks. It contains three real-world sub-datasets, Web-bird, Web-aircraft, and Web-car. They utilize the fine-grained category labels of benchmark datasets CUB200-2011\upcite{wah2011caltech}, FGVC-aircraft\upcite{aircraft}, and Cars-196\upcite{car196} as target categories to crawl web images from Bing Image Search Engine. Compared with benchmark datasets, they have a larger number of training samples (18388, 13503, and 21448, respectively) but potential noisy labels. The test sets are directly taken from three benchmark datasets (CUB200-2011, FGVC-aircraft, and Stanford Cars).
	Details of three sub-datasets are as follows and summarized in Table \ref{WebFG-496}.
	
	\textbf{Web-bird} consists of 200 different subcategories of birds. Its web training set contains 18388 images and the test set has 5794 clean samples. The training label accuracy estimated by random sampling is about 65\%.
	
	\textbf{Web-aircraft} covers 100 variants of aircraft. The number of web training samples and clean test images are 13503 and 3333, respectively. Its estimated web label accuracy is approximately 73\%.
	
	\textbf{Web-car} contains 196 types of vehicles. Its training set consists of 21448 web samples with a roughly estimated label accuracy of 67\%. The test set contains 8041 manually labeled images.
	
	We follow the evaluation metric in section~\ref{cifar_setting} and utilize ACA to evaluate model performance.

	\subsection{Implementation Details} 
	We adopt a ImageNet pre-trained ResNet-50\upcite{he2016deep} model as our backbone network and resize input images to $ (448,448) $ with random horizontal flip as weak data augmentation. The network is trained for $ 100 $ epochs with batch size set to $ 32 $. The learning rate is initialized to $ 0.01 $ and decreases in a cosine annealing manner\upcite{loshchilov2016sgdr}.
	The momentum and weight decay of stochastic gradient descent (SGD)\upcite{bottou2010large} optimizer are set to $ 0.9 $ and $ 10^{-5} $, respectively.
	
	As for hyper-parameters, we set momentum $ m $ for EMA and weight $ w $, $ \gamma $ to $ 0.5 $. Warm-up epoch $ t_{m} $ is empirically set to $ 5 $. We set $ \alpha $ to $ 0.5 $ on Web-bird and adopt a linear decreasing $ \alpha $ on Web-aircraft and Web-car. Specifically, $ \alpha $ linearly decreases from $ 1 $ to $ 0.8 $ and $ 0.3 $ in $ 5 $ epochs on Web-aircraft and Web-car, respectively, which is motivated by the increasing drop rate trick\upcite{han2018co}. Threshold $ \tau $ for relabeling is set to $ 0.04, 0.02, 0.04 $ on Web-bird, Web-aircraft, and Web-car, respectively.
	Our experiments are all performed on one NVIDIA Tesla V100 GPU.

	\subsection{Baseline Methods}
	On real-world datasets, our baselines contain the following state-of-the-art methods: Decoupling\upcite{malach2017decoupling}, Co-teaching\upcite{han2018co}, Sub-center\upcite{deng2020sub}, Co-teaching+\upcite{yu2019does}, Self-Adaptive\upcite{huang2020self}, JoCoR\upcite{wei2020combating}, DivideMix\upcite{li2019dividemix},	Jo-SRC\upcite{zhang2021delving},	OLS\upcite{zhang2021delving}, PLC\upcite{zhang2021learning}, Peer-learning\upcite{sun2021webly}, SELC\upcite{selc}, Co-LDL\upcite{coldl}, AGCE\upcite{asymloss} and CMW-Net-SL\upcite{cmwnet}. We reproduce most of the above baselines\upcite{malach2017decoupling,han2018co,deng2020sub,yu2019does,huang2020self,wei2020combating,li2019dividemix,zhang2021delving,zhang2021learning,sun2021webly,selc,asymloss} using the same backbone network for fair comparisons. In addition, we also train a ResNet-50 network using the cross-entropy loss function for comparison (Backbone).

			\begin{figure*}[t]
	\includegraphics[width=\linewidth]{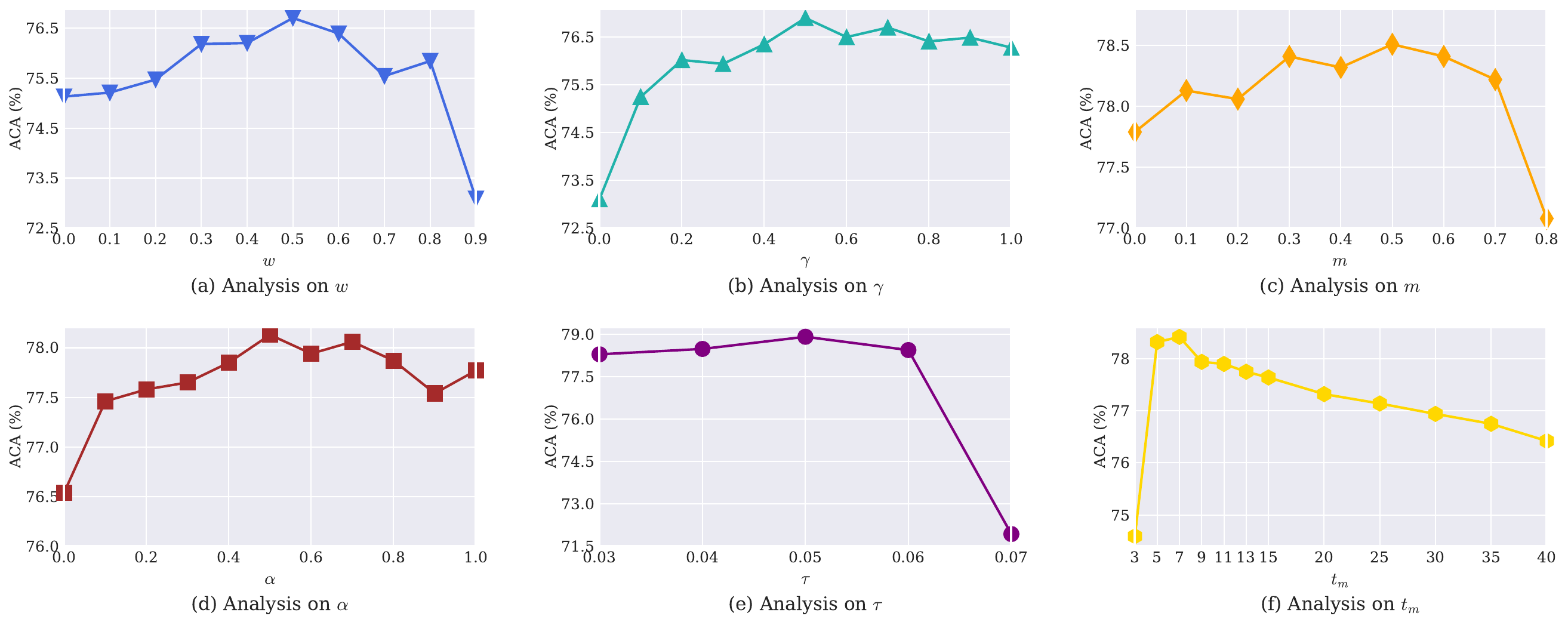}
	\caption{The parameter sensitivities of on Web-bird with a ResNet-18 as the backbone.}
	\label{parameter}
\end{figure*}

	\subsection{Experimental Results and Analysis}

	We demonstrate ACA performances of baseline approaches and GRIP on WebFG-496 in Table \ref{tab1}. From Table \ref{tab1}, we can observe that GRIP significantly outperforms baselines on all three datasets. Specifically, it surpasses the best results of baselines by $ 1.01 \% $, $ 0.56 \% $, and $ 1.36 \% $ on Web-bird, Web-aircraft, Web-car, respectively. For clearer illustration, we also present test accuracy trending of GRIP and compare it with some representative noise-cleaning baselines in Fig.~\ref{aca}. As illustrated in Fig.~\ref{aca}, GRIP shows higher accuracies and faster training speeds than other noise-cleaning approaches\upcite{malach2017decoupling,wei2020combating} on all benchmark datasets. This superiority owes to our group regularization strategy, which significantly improves the noise-robustness. In addition, compared with noise-robust approaches\upcite{deng2020sub,zhang2021delving,selc,asymloss,cmwnet} in Table \ref{tab1}, our approach achieves higher test accuracies by leveraging the instance purification strategy to specifically tackling noisy labels. The significant improvements over baselines demonstrate the effectiveness of simultaneously leveraging both noise-robust and noise-cleaning strategies for combating noisy labels.
	
	Particularly, GRIP shows superior performance than DivideMix\upcite{li2019dividemix}, which utilizes noise-cleaning and noise-robust algorithms simultaneously. DivideMix has large gaps to GRIP on Web-bird ($ 8.13 \% $) and Web-car ($ 5.22 \% $). The reason may lie in the Gaussian Mixture Model (GMM)\upcite{permuter2006study} used for dividing training samples in DivideMix. It can be difficult to fit a GMM on the real-world noisy data distribution. If GMM fails, the performance will inevitably decline.
	
	Furthermore, we can observe from Table \ref{tab1} that some baselines (Decoupling\upcite{malach2017decoupling}, Co-teaching+\upcite{yu2019does}, Self-Adaptive\upcite{huang2020self}, PLC\upcite{zhang2021learning}, CMW-Net-SL\upcite{cmwnet}) only show slight improvements or even inferior performances to Backbone. The reason can be that they are designed and tested on coarse-grained synthetic noisy datasets. As a result, they tend to be less practical for fine-grained tasks in a real-world scenario. Compared with them, our approach is more effective in practical application.

	\section{Ablation Studies}
	
	In order to further analyze our approach, we conduct experiments on the real-world dataset Web-bird using ResNet-18 by default.
	
	\subsection{Parameter Analysis}
	
	In this experiment, we investigate the parameter sensitivities of weights $ w $ and $ \gamma $ for loss functions, momentum $ m $ for EMA, $ \alpha $ and $ \tau $ for noise-cleaning, and warm-up epoch $ t_{m} $. Although our approach seems to have many parameters, we will show that half of them are robust and easy to adjust. The experimental results are shown in Fig.~\ref{parameter}. 
	
	We set $ \gamma $ to $ 0.5 $ and changes $ w $ from $ 0 $ to $ 0.9 $ for investigation. From Fig.~\ref{parameter}~(a), we can observe that the performance first steadily increases to the optimal value as $ w $ rises. This phenomenon indicates that $ \mathcal{L}_{soft} $ boosts the robustness with a proper $ w $. However, if $ w $ further increases, the supervision of $ \mathcal{L}_{CE} $ is weakened, which results in a performance decline. To achieve the best performance, we set $ w $ to $ 0.5 $ on all datasets, which is a balanced weight between $ \mathcal{L}_{soft} $ and  $ \mathcal{L}_{CE} $. In addition, we can also see that $ w $ is robust in $ [0.3, 0.6] $, which indicates a relatively balanced $ w $ can work well.
	
	Similar to the analysis on $ w $, we set $ w $ to $ 0.5 $ and changes $ \gamma $ from $ 0 $ to $ 1 $. We can observe from Fig.~\ref{parameter}~(b) that the performance climbs as $ \gamma $ rises from $ 0 $ to $ 0.5 $, then slightly decreases when $ \gamma $ further increases. Supported by this result, we simply set $ \gamma $ to $ 0.5 $ on all datasets. We can also find that $ \gamma $ is robust in $ [0.4, 1] $. 
	
	We analyze the effect of $ m $ in Fig.~\ref{parameter}~(c). It can be observed from Fig.~\ref{parameter}~(c) that unless $ m $ is too large (over 0.8), applying EMA can boost the performance over the baseline ($ m $=$ 0 $). Furthermore, EMA can work well and achieve close test accuracies in $ [0.3, 0.7] $.
	
	We do not perform label re-assignment and only analyze the effect of $ \alpha $ on sample selection in Fig.~\ref{parameter}~(d). When $ \alpha $ increases from $ 0 $ to $ 0.5 $, the performance steadily climbs because more training samples are utilized. However, when $ \alpha $ is larger than $ 0.5 $, the performance declines with some fluctuations. The reason is that fewer noisy images are discarded and thus noise-cleaning becomes less effective.
	
	\begin{figure}[t]
	\includegraphics[width=0.47\textwidth]{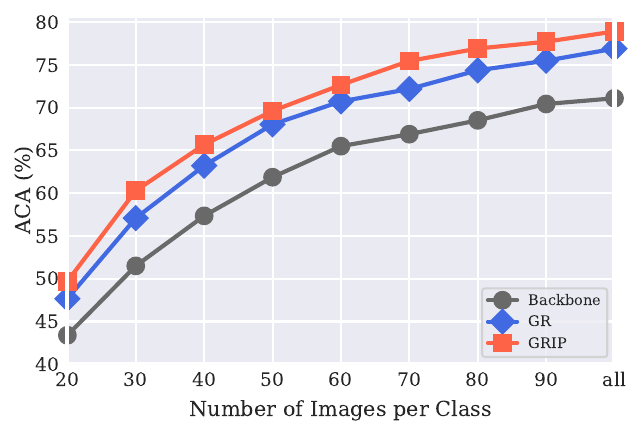}
	\centering
	\caption{ACA (\%) vs. number of images per class. GR: Group Regularization; GRIP: Group Regularization and Instance Purification. "all" indicates using the entire dataset.}
	\label{numbers}
\end{figure}

	The analysis of $ \tau $ is illustrated in Fig.~\ref{parameter}~(e). From Fig.~\ref{parameter}~(e), we can see that performance climbs as $ \tau $ increases from $ 0.03 $ to $ 0.05 $. Then it drops fast when $ \tau $ becomes larger ($ \tau=0.07 $). The reason can be that too many noisy samples are relabeled and reused. Some of them may still have false labels or even be out-of-distribution (OOD). From this result, we believe that a small value of $ \tau $ is safe and can boost performance.
	
	The analysis of $ t_{m} $ is illustrated in Fig.~\ref{parameter}~(f). We can observe from Fig.~\ref{parameter}~(f) that if the warm-up period is too short (less than $ 3 $), the model cannot learn reliable soft labels for noise identification, which results in an unsatisfying performance. A proper $ t_{m} $ lies in $ [5, 10] $. If $ t_{m} $ further increases, the model will be affected by noisy labels in the warm-up period and noise-cleaning becomes less useful, resulting in declining test accuracies.
	
	From the experimental results in Fig.~\ref{parameter}, we can conclude that weights $ w $, $ \gamma $, and momentum $ m $ are robust and easy to adjust. Parameters $ \alpha $ and $ \tau $ for noise-cleaning should be changed on each dataset because they are concerned with noise ratios. As for $ t_{m} $, a relatively small value is recommended, \eg in $ [5, 10] $.

	\subsection{Analysis on Dataset Sizes}
	
	In this experiment, we investigate the applicability of GRIP on small datasets by changing the number of web images used for each category on Web-bird\upcite{sun2021webly}. To be detailed, we construct sub-datasets from Web-bird by randomly selecting $ 10 $ to $ 90 $ samples for each class. Then we compare the performance of the backbone network, our group regularization strategy (GR), and our proposed GRIP on each sub-dataset in Fig.~\ref{numbers}. It can be observed from Fig.~\ref{numbers} that our group regularization strategy shows significant and consistent improvements across different dataset sizes over the baseline. After applying the instance purification strategy, GRIP further boosts the performance on each sub-dataset. It shows remarkable noise-robustness even when the dataset is rather small, \eg $ 20 $ images per class. From the experimental results, we can conclude that our approach is insensitive to the dataset size.
	
	We can also see from Fig.~\ref{numbers} that the ACA performance climbs steadily when more training samples are utilized. Therefore, leveraging free web images is a promising research direction as it allows boosting model performance and robustness through enlarging datasets.
	
	
	\begin{table}[tb]
		\centering
		\renewcommand{\arraystretch}{1.1}
		\caption{The ACA (\%) performances and improvements of different backbones on Web-bird. Baseline indicates that the network is straightforwardly trained using cross-entropy loss.}		
		\setlength{\tabcolsep}{5pt}
		\scalebox{0.9}{
			\begin{tabular}{c|c|c|c} 
				\hline 
				\textbf{\:\:\:Backbone\:\:\:} & \textbf{\:\:\: Method \:\:\:} & \textbf{Performance} & \textbf{Improvement}   \\
				\hline
				\multirow{2}{*}{VGG-16}	
				& Baseline 	&    66.34        	&  
				\multirow{2}{*}{$\Delta$ 9.53} \\
				& Ours	    &  	  \textbf{75.87}         &  \\
				\hline
				\multirow{2}{*}{ResNet-18}			
				& Baseline &      71.10        	& 
				\multirow{2}{*}{$\Delta$ 7.81} \\ 
				& Ours     &  	  \textbf{78.91}         & \\
				\hline
				\multirow{2}{*}{ResNet-50}	 
				& Baseline &	  73.30				& 
				\multirow{2}{*}{$\Delta$ 9.23} \\ 
				& Ours     &	\textbf{82.53}  			& \\
				\hline
		\end{tabular}}
		\label{tab4}
	\end{table}
	
	\subsection{Applicability Across Different Backbones}
	We test our approach using different backbone networks on Web-bird to analyze the applicability in Table \ref{tab4}. We can observe from Table \ref{tab4} that our approach boosts the performance by $ 9.53 \% $, $ 7.81 \% $, and $ 9.23 \% $ on VGG-16\upcite{simonyan2014very}, ResNet-18, and ResNet-50, respectively. The experimental results indicate that our approach is robust and shows remarkable improvements across different CNN architectures.
	
	\subsection{Contribution of Each Component}
	In this subsection, we gradually add components in our method to the baseline model and present the contribution of each component in Table \ref{tab2}. From lines  $ 1 $ to $ 6 $ in Table \ref{tab2}, we can observe that all components contribute to the performance improvements. To be detailed, leveraging $ \mathcal{L}_{Soft} $ surpasses the baseline by around $ 2 \% $. Then we further makes remarkable improvements through applying $ \mathcal{L}_{ME} $ ($ 3.35 \% $) and EMA ($ 0.44 \% $). Owing to our group regularization strategy, the ACA performance reaches $ 76.9 \% $ and significantly surpasses the baseline. After further applying instance purification, the final performance reaches $ 78.91 \% $ by employing JS divergence noise identification ($ 1.42 \% $) and relabeling ($ 0.59 \% $). Resorting to the contribution of each component, our approach boosts the final performance by $ 7.81 \% $ over the baseline.

	\begin{table*}[t]
		\centering
		\renewcommand{\arraystretch}{1.1}
		\caption{The contribution of each component in our approach ($ 1 $ to $ 6 $) and comparisons with other similar methods ($ 7 $ to $ 11 $). The abbreviations are as follows: 
			$ \mathcal{L}_{Soft} $ (Soft), $ \mathcal{L}_{ME} $ (ME), exponential moving average (EMA), JS divergence noise identification criterion (JS), relabeling (RE), label smoothing (LS), small-loss principle (SL) and JS divergence noise identification within each mini-batch (MB). `$ \checkmark $' indicates the component is utilized in training.}
		\setlength{\tabcolsep}{15pt}
		\scalebox{0.97}{
			\begin{tabular}{c|c|c|c|c|c|c|c|c|c}
				\hline
				\multirow{2}{*}{\textbf{No}} & \multicolumn{3}{c|}{\textbf{Regularization}} & \multicolumn{2}{c|}{\textbf{Purification}} & \multicolumn{3}{c|}{\textbf{Comparisons}} & \multirow{2}{*}{\textbf{ACA}} \\
				\cline{2-9} &  Soft  & ME & EMA & JS & RE & LS  & SL & MB  \\		
				\hline
				1 &	 & 	& 	& 	& 	& 	& 	& 	& 71.10 \\
				2 &	\checkmark & 	& 	& 	& 	& 	& 	& 	& 73.11 \\
				3 &	\checkmark &  \checkmark	& 	& 	& 	& 	& 	& 	& 76.46 \\
				4 &	\checkmark &  \checkmark	&  \checkmark	& 	& 	& 	& 	& 	& 76.90 \\
				5 &	\checkmark &  \checkmark	&  \checkmark	& \checkmark & 	& 	& 	& 	& 78.32 \\
				6 &	\checkmark &  \checkmark	&  \checkmark	& \checkmark & \checkmark	& 	& 	& 	& \textbf{78.91} \\
				\hline
				7 &	 &  \checkmark	&  	&  & 	& 	& 	& 	& 75.13 \\
				8 &	 &  	&  	&  & 	& \checkmark	& 	& 	& 72.28 \\
				9 &	 &  \checkmark	&  	&  & 	& \checkmark	& 	& 	& 75.18 \\
				10 &	\checkmark &  \checkmark	&  \checkmark	&  & 	& 	& \checkmark	& 	& 77.58 \\
				11 &	\checkmark &  \checkmark	&  \checkmark	&  & 	& 	& 	& \checkmark	& 77.98 \\
				\hline
		\end{tabular}}
		\label{tab2}
	\end{table*}
	
	\begin{figure*}[t]
		\includegraphics[width=0.99\textwidth]{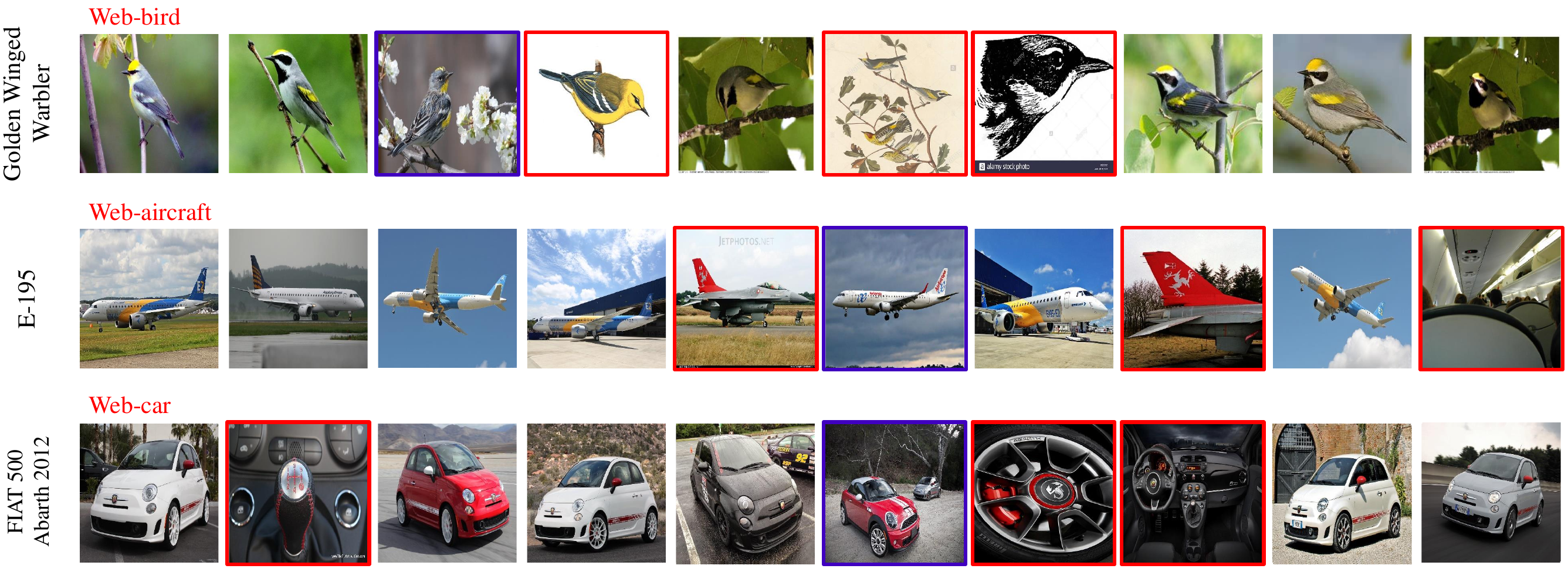}
		\caption{Illustration of noise-cleaning results on Web-bird, Web-car, and Web-aircraft. Each row illustrates ten samples that are from the same fine-grained category. Blue and red boxes indicate revisable and discarded samples, respectively.}
		\label{noise-cleaning}
	\end{figure*}
	
	\subsection{Comparisons}
	In this experiment, we demonstrate the superiority of our proposed approach over other similar strategies. We first compare the proposed group regularization with the widely-used LS trick. Then we compare JS divergence noise identification with the small-loss principle. We also compare the global and mini-batch selection based on our JS divergence principle.
	
	The experimental results are presented in Table \ref{tab2} (lines  $ 7 $ to $ 11 $). We can observe from Table \ref{tab2} that LS shows a slightly inferior performance to simply leveraging soft label supervision $ \mathcal{L}_{Soft} $. Applying LS and maximum entropy strategy $ \mathcal{L}_{ME} $ together shows nearly no improvement over utilizing $ \mathcal{L}_{ME} $ alone. On the contrary, combining $ \mathcal{L}_{Soft} $ and $ \mathcal{L}_{ME} $ boosts the performance significantly. This result demonstrates the advantages of utilizing estimated soft labels over LS: higher performance and better flexibility. Owing to $ \mathcal{L}_{Soft} $ and $ \mathcal{L}_{ME} $, our group regularization strategy surpasses LS remarkably.

	Comparing lines $ 10 $ and $ 11 $ in Table \ref{tab2}, we can find that noise identification within mini-batch based on JS divergence is superior to that based on loss. This result supports our argument that utilizing class soft labels is more effective than simply relying on loss values in noise identification. Comparing ling $ 5 $ and $ 11 $, we can observe that global selection further boosts the performance. The improvement derives from that the noise rate imbalance problem is alleviated.
	
	Note that we apply $ \mathcal{L}_{ME} $ on discarded images. In order to demonstrate the contribution of this design, we remove it for comparison. Then we find that the performance drops from $ 78.32 \% $ to $ 77.66 \% $. Since noisy samples are utilized for training in the warm-up period, the network memorizes them to some extent. They potentially misguide noise identification and degrade performance. To solve this problem, $ \mathcal{L}_{ME} $ provides a strong regularization to guide the network to forget noisy samples and make them farther from class soft labels to guarantee more reliable noise identification.
	
	\subsection{Noise-cleaning Visualization}
	We sampled noise-cleaning results on WebFG-496 and visualize them in Fig.~\ref{noise-cleaning}. We can observe from Fig.~\ref{noise-cleaning} that our method can effectively divide clean, revisable, and discarded samples. We can also notice that web dataset inevitably contain noisy labels. For example, searching web images for cars has a risk of getting some images of tires and steering wheels. This phenomenon reminds us that noise-cleaning operation is necessary for training robust models in learning with noisy labels tasks.
	
	\section{Conclusion}
	In this paper, we proposed an effective training method named GRIP that leverages group regularization to benefit instance purification on both synthetic and real-world datasets. The proposed group regularization strategy generates reliable class soft labels to boost model robustness against label noise.
	By measuring the differences between instances and class soft labels, our method can globally identify noisy and revisable samples. Resorting to the regularization from the category aspect and purification at the instance level, GRIP inherits the advantages of both noise-robust and noise-cleaning strategies. By conducting comprehensive experiments, we demonstrate the superiority of our approach over existing methods in combating noisy labels on both synthetic and real-world datasets.

\bibliographystyle{bib/IEEEtran}
\bibliography{bib/cs_bib}

\end{document}